\documentclass[]{spie}  %>>> use for US letter paper
\usepackage{amsmath,amsfonts,amssymb}
\usepackage[pdftex]{graphicx}
\usepackage{hyperref}
\usepackage{booktabs}
\usepackage{multirow}
\usepackage{float}
\usepackage{xcolor}
\usepackage[caption=false,font=normalsize,labelfont=sf,textfont=sf]{subfig}

\DeclareGraphicsExtensions{.pdf,.jpeg,.png,.jpg}

\title{Combining Visible and Infrared Spectrum Imagery using Machine Learning for Small Unmanned Aerial System Detection}

\author[a]{Vinicius G. Goecks}
\author[b]{Grayson Woods}
\author[a]{John Valasek}
\affil[a]{Department of Aerospace Engineering, Texas A\&M University}
\affil[b]{Department of Mechanical Engineering, Texas A\&M University}

\authorinfo{Further author information:\\
Vinicius G. Goecks, Graduate Research Assistant, Vehicle Systems $\&$ Control Laboratory, vinicius.goecks@tamu.edu\\
Grayson Woods, Graduate Research Assistant, Unmanned Systems Laboratory, graysonwoods@tamu.edu\\
Niladri Das, Graduate Research Assistant, Intelligent Systems Research Laboratory, niladridas@tamu.edu\\
John Valasek, Professor and Director, Vehicle Systems $\&$ Control Laboratory, valasek@tamu.edu\\}

\begin{document} 
\maketitle

\begin{abstract}

There is an increasing demand for technology and solutions to counter commercial, off-the-shelf small unmanned aerial systems (sUAS).
Advances in machine learning and deep neural networks for object detection, coupled with lower cost and power requirements of cameras, led to promising vision-based solutions for sUAS detection.
However, solely relying on the visible spectrum has previously led to reliability issues in low contrast scenarios such as sUAS flying below the treeline and against bright sources of light.
Alternatively, due to the relatively high heat signatures emitted from sUAS during flight, a long-wave infrared (LWIR) sensor is able to produce images that clearly contrast the sUAS from its background.
However, compared to widely available visible spectrum sensors, LWIR sensors have lower resolution and may produce more false positives when exposed to birds or other heat sources.
This research work proposes combining the advantages of the LWIR and visible spectrum sensors using machine learning for vision-based detection of sUAS. Utilizing the heightened background contrast from the LWIR sensor combined and synchronized with the relatively increased resolution of the visible spectrum sensor, a deep learning model was trained to detect the sUAS through previously difficult environments. More specifically, the approach demonstrated effective detection of multiple sUAS flying above and below the treeline, in the presence of heat sources, and glare from the sun.
Our approach achieved a detection rate of 71.2 $\pm$ 8.3\%, improving by 69\% when compared to LWIR and by 30.4\% when visible spectrum alone, and achieved false alarm rate of 2.7 $\pm$ 2.6\%, decreasing by 74.1\% and by 47.1\% when compared to LWIR and visible spectrum alone, respectively, on average, for single and multiple drone scenarios, controlled for the same confidence metric of the machine learning object detector of at least 50\%.
With a network of these small and affordable sensors, one can accurately estimate the 3D position of the sUAS, which could then be used for elimination or further localization from more narrow sensors, like a fire-control radar (FCR).
Videos of the solution's performance can be seen at \url{https://sites.google.com/view/tamudrone-spie2020/}.

% SUMMARY (100 Words): There is an increasing demand for technology and solutions to counter commercial, off-the-shelf small unmanned aerial systems (sUAS) that pose a potential threat when used by harmful, unknown agents. The proposed solution from this paper combined LWIR and visible spectrum imagery using machine learning for vision-based detection of sUAS. Utilizing both the heightened background contrast from the LWIR sensor and the relatively increased resolution of the visible spectrum sensor, a deep learning model was trained to detect sUAS in previously difficult environments (e.g. above and below the treeline, in the presence of birds and glare from the sun).
\end{abstract}

% Include a list of keywords after the abstract 

% ----------------------------------------------------
% INTRODUCTION
% ----------------------------------------------------
\section{INTRODUCTION}
\label{sec:intro}  % \label{} allows reference to this section

% context for your ideas
% \comment{talks about how easy it is to acquire drones and that they can be used for bad things. To prevent these bad things people have proposed counter-drone solutions. One of the promising ones are vision-based because cameras are cheap and there was a lot of research advancement on neural networks for object detection that could be used to detect drones.}

Small unmanned aircraft systems (sUAS), commonly called \textit{drones}, were once only thought of as military aircraft. 
We have seen images from the government or the news media showing comparatively larger aircraft such as the General Atomics MQ-9 Reaper, which are being controlled by an operator sitting hundreds of miles away. 
They are effective in performing surveillance but can also deploy weapons without risking its operator. 
Increasingly, the technology to control and operate such unmanned aircraft is becoming affordable and easily available. 
We are seeing a rise in the manufacturing of consumer-grade and commercial-grade sUAS.
Companies such as DJI\cite{DJI}, Parrot\cite{PARROT}, and Autel Robotics\cite{AUTEL} have drones equipped with high definition RGB and infrared cameras, sophisticated mission planners, and are capable of completing its mission autonomously, that can be bought for as little as $\$$1000 USD.
% Some retails companies have started exploiting this easy availability and reliability of drones.
% Companies such as Amazon are seriously considering business plans that utilize drones for commercial endeavors.
% It is also becoming easier to assemble a drone with off-the-shelf parts which can be bought online.
This capability is posing crucial privacy and security threats to people and organizations around the world, specially when these drones are retrofitted with sophisticated cameras and possibly weapons, which sparked a need for counter-sUAS, or counter-drone, solutions.

One of the components of a counter-drone solution is the \textit{detection} of the drone.
Due to the wide availability of cameras and advancements in deep learning and computer vision techniques, vision-based solutions became more appealing to address the detection problem.
Vision based-object detection techniques were mainly developed to detect objects on the visible spectrum while using RGB cameras.
However, accuracy strongly decreases when detecting objects under fog, rain, or in poorly illuminated environments.
% In situation where it is critical to have zero miss detection investigating diverse imaging techniques is justified.
As an alternative, long-wave infrared (LWIR) cameras are more robust to these conditions but are less accurate due to lower resolution and false positives when detecting objects with similar heat signatures such as differentiating at long distance drones from birds or light bulbs.

% proposed solution or response too the problem • This is your main point and main claim
To leverage the advantages of each sensor, our proposed solution combines images captured from RGB and LWIR cameras to train a machine learning model for the detection of sUAS. Our main contribution is the insight of utilizing the heightened background contrast from the LWIR sensor combined and synchronized with the relatively increased resolution of the RGB sensor to train a deep learning model to detect sUAS through previously difficult environments. More specifically, the approach demonstrated effective detection of multiple sUAS flying above and below the treeline, in the presence of heat sources, and glare from the sun.

% ----------------------------------------------------
% RELATED WORK
% ----------------------------------------------------
\section{Related Work}

For large unmanned aerial systems (UAS) detection, the majority of techniques use radar imaging \cite{moffatt1975detection,bullard1991pulse,khan1994target,khan1995aircraft,headrick1998applications,brown2010air}. However, the size and speed of small unmanned aerial systems (sUAS) present challenges for radar detection, which must be able to scan a large area with discrimination to prevent false alarms from birds and moving ground objects \cite{poitevin2017challenges}. Deep learning techniques have been developed to help detect and classify sUAS by recognizing patterns in the Doppler signature of micro UAS \cite{mendis2016deep}. However, the confidence with which these systems can be practically employed for counter-sUAS is still up for debate.

Image-based sUAS detection provides another solution for the detection of sUAS.
Visual sensors have been used to track targets moving over complex backgrounds \cite{rozantsev2016detecting}. In addition, temporal information from subsequent images can be supplemented to make the tracking robust to when the background or target object is changing in appearance \cite{aker2017using}. Detection of sUAS with visual sensors can even be employed when the sensors themselves are moving (for example, when onboard another sUAS \cite{opromolla2018vision}). Image processing techniques such as feature detection and corner detection can also be used to analyze an object or scene from two different positions and find features that would correspond to a real-world 3D object or scene \cite{rosten2006machine}, offering an increased ability to classify the target object.

Fusing multi-modal sensor outputs such as RGB and infrared video streams are gaining prominence to detect objects under changing environmental conditions. If one sensing modality is impaired in a particular condition, other sensing modalities might not be affected. The fusion of sensor data for object detection is typically done in any of these three phases: at the pixel level, at the feature level, or at the decision level. RGB and infrared sensor data fused at a pixel level \cite{St_Laurent_2007} can then be segmented using either parametric \cite{Stauffer}techniques such as Gaussian Mixture Model or non-parametric techniques such as kernel-based density estimators. A new non-parametric technique based on codebook model \cite{Kim_2005} has shown enhanced object detection performance \cite{St_Laurent_2007} when applied to pixel-level fused data. 
The image segmentation accuracy as measured using the \textit{detection rate} metric is considerably higher for the fused data compared to RGB and infrared data separately\cite{St_Laurent_2007}. Moreover the processing time for fused data was found to be less compared to the summation of the processing time required for RGB and infrared data individually. The detection accuracy is shown to be further improved if we allow a context-aware, quality-based fusion \cite{Alldieck_2016}. Context sensitive indicators are introduced \cite{Alldieck_2016} to weigh in the soft segmentation performed over individual RGB and infrared data.

More recently, machine learning approaches were developed to remove the necessity of hand-designing image features and corners to, instead, learn these patterns automatically from large datasets.
Simple Online and Realtime Tracking with a Deep Association Metric (Deep SORT) \cite{Bewley2016_sort,wojke2017simple} trained a convolutional neural network on a large-scale person identification dataset to learn an association metric for each target identified. Paired with Kalman filtering in the image-space and frame-by-frame data association, this approach achieves favorable performance when tracking objects for longer periods of time under occlusion.
One of the most common approaches, You Only Look Once (YOLO) \cite{redmon2013darknet,redmon2016you,redmon2018yolov3}, frames object detection as a regression problem where a neural network predicts, in real time, bounding boxes and class probabilities directly from full images in only one evaluation step.
Improving upon YOLO, EfficientNet \cite{mingxing2019efficientnet} proposes the compound scaling method that uniformly scales depth, width, and resolution of convolution neural networks following a fixed ratio to achieve state-of-the-art object detection while using smaller and faster networks. 
Combined with machine learning, data from visual sensors can be used to not only track sUAS \cite{wang2013learning} but also distinguish them from birds \cite{aker2017using} and other background objects.
Our approach, detailed below, utilizes many of these advancements to provide a robust framework for sUAS detection.

% ----------------------------------------------------
% METHODS
% ----------------------------------------------------
\section{Methods}

\begin{figure*}[!htb]%
    \centering
    \subfloat[]{{\includegraphics[width=0.3\linewidth]{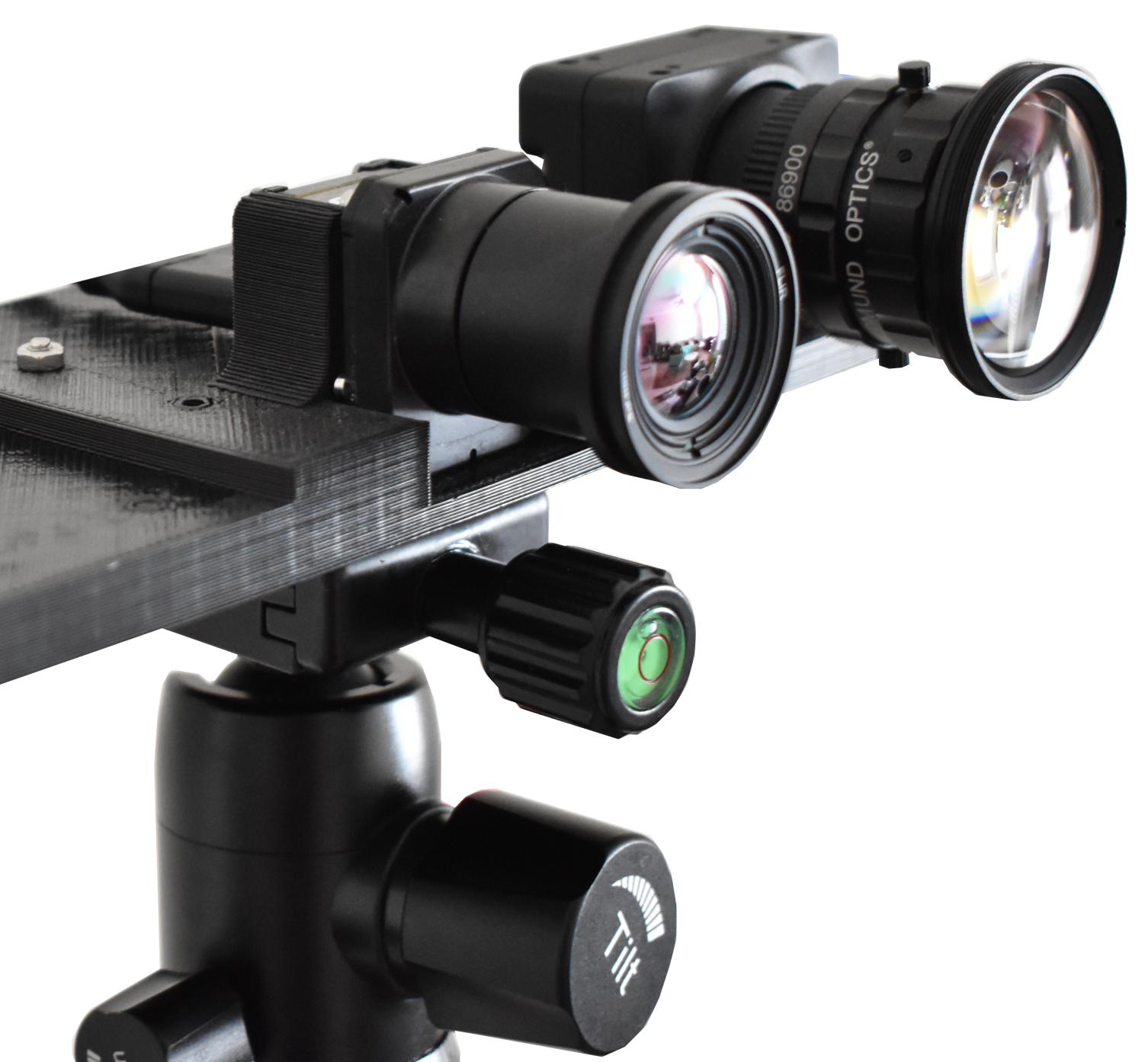}\label{fig:camera_setup_a} }}%
    \subfloat[]{{\includegraphics[width=0.67\linewidth]{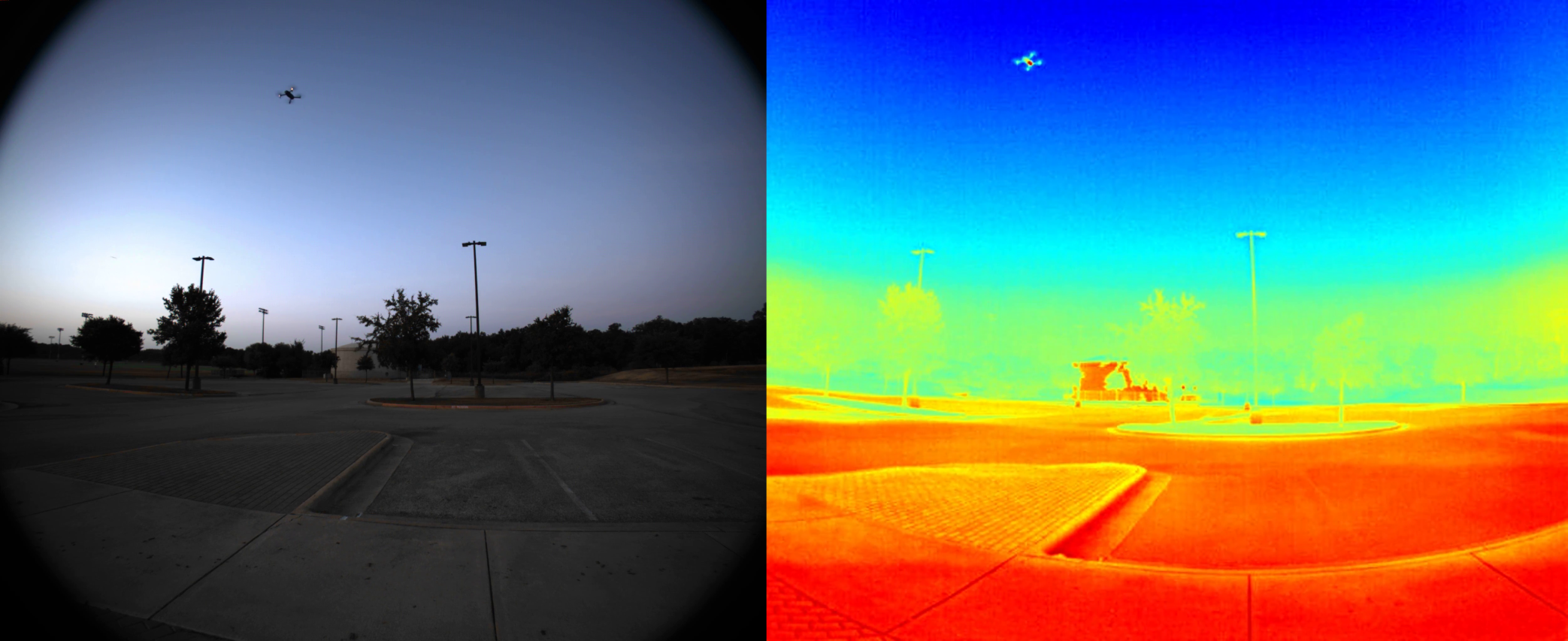}\label{fig:camera_setup_b} }}
    \caption{(a) Picture of the mounted RGB and LWIR cameras used to capture data for this research work and (b) sample RGB and its respective LWIR frame after synchronization and before fusion.}%
    \label{fig:camera_setup}%
\end{figure*}

This research work proposes combining the advantages of the LWIR and visible spectrum sensors using machine learning for vision-based detection of sUAS.
Due to the power and battery discharge rates required to fly small unmanned quadrotors, these vehicles hold a characteristic heat signature when imaged with LWIR sensors that contrasts with common cluttered backgrounds suck as buildings, trees, sky, and the terrain.
The same does not happen in the visible spectrum where changes in lighting and cluttered background can completely camouflage a flying small unmanned quadrotor.
Conversely, visible spectrum sensors are widely available in the market and have higher resolution, which aids detection at long-ranges assuming that the vehicle contrasts with the background.
Our prototype, as seen in Figure \ref{fig:camera_setup_a} consists of two calibrated cameras aligned based on the position of their image capture sensor and attached to a custom mount system that allows them to be attached to a camera tripod.
The cameras, from left to right, are a FLIR infrared camera BOSON 640 coupled with 4.9mm lens and a FLIR RGB camera CM3-U3-50S5C-CS Chameleon3 USB3 coupled with a 3.5mm C Series fixed focal length lens. The camera specifications are shown in Table \ref{tab:camera_info}.
All camera data was streamed and recorded using ROS (Robot Operating System) \cite{quigley2009ros}, synchronized to run at $30$ Hz.
A sample RGB and respective LWIR frame demonstrating the data capture system can be seen in Figure \ref{fig:camera_setup_b}.

\begin{table}[!htbp]
\centering
\caption{Camera Specifications.} 
\label{tab:camera_info} 
\begin{tabular}{@{}lcc@{}}\toprule%[1.25pt]
 & \multicolumn{2}{c}{\textbf{Camera Type}} \\ \cmidrule(l){2-3} 
\textbf{Specifications} &  RGB & Infrared \\ \midrule
Model & FLIR CM3-U3-50S5C-CS & FLIR BOSON 640 \\
Lens focal length (mm) & 3.5 & 4.9 \\
Frame Rate (Hz) & $35$ & $60$ \\
Resolution (pixels) &  $2448 \times 2048$ & $640 \times 512$ \\
Sensor Type & CMOS & N/A \\
Sensor Format & $2/3"$ & N/A \\
Thermal Sensitivity (mK) & N/A & $< 60$ \\
\bottomrule%[1.25pt]
\end{tabular}
\end{table}

% \comment{Industrial classification of our prototype:}The current prototype has been tested and approved on a relevant environment, as seen on this video, classifying our approach as Technology Readiness Level 6 (TRL).  Since our prototype has only been manufactured in a laboratory environment our solution classifies as MRL 4 in terms of Manufacturing Readiness Levels (MRL),. The prototype can also be classified as vehicle-mounted (1-man lift) since it exceeds 5 pounds in weight which includes the host laptop. The host laptop can be powered by either a 12 volts DC vehicle plug coupled with a car inverter or the self-contained laptop batteries. The system can also be scaled down to be “man-portable” and run off body worn power source, \textcolor{red}{such as 2590 batteries} by replacing the laptop by a single-board computer (for example, NVIDIA TX2 or XAVIER) and smaller screen.

The data collection process consisted of imaging two different sUAS, DJI Mavic Pro and DJI Phantom 4, while flying at different times of the day and weather conditions, such as during sunset and sunrise in a clear sky or during the afternoon of a cloudy day, and under different lighting conditions in order to create a diverse dataset to be used as training data for the machine learning model.
A total of 30 minutes of data at 30 frames per second were recorded for a total of 52800 frames for each camera, LWIR and RGB. From that, 1275 frames in total were manually labeled by drawing a bounding box surrounding the vehicle.
The main assumption during this process was that the person should label LWIR and RGB data independently so there is no information leakage between one spectrum to another.

\begin{figure}[!htb]
    \centering
    \includegraphics[width=0.985\linewidth]{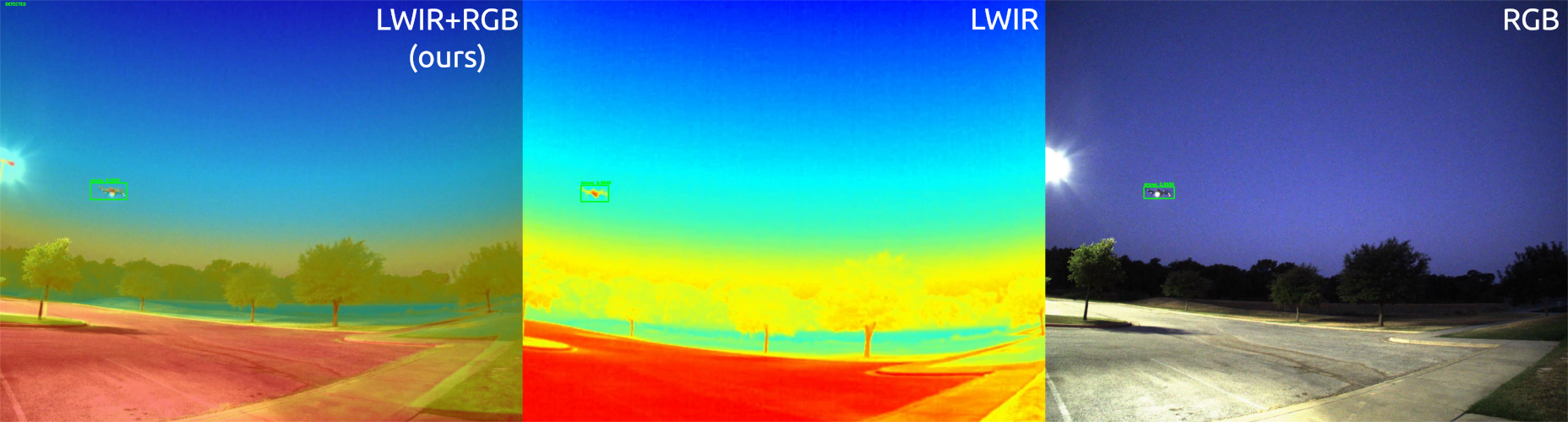}
    \caption{Sample of frames illustrating our proposed approach (LWIR+RGB, leftmost image) compared to LWIR and RGB alone to detect sUAS. The proposed method synchronizes and combines both LWIR and RGB images leveraging the advantages of both sensors for object detection.}
    \label{fig:sample_blended}
\end{figure}

To combine the calibrated and synchronized LWIR and RGB images, we first computed the homography transformation from the LWIR to the RGB image frame so each pixel in the infrared spectrum could be associated with one pixel in the visible spectrum.
The pixels were combined by computing the equally weighted sum between them so the resulting pixel carries information from both spectra.
Figure \ref{fig:sample_blended} illustrates our proposed approach to combined LWIR+RGB data in comparison to LWIR and RGB alone.
The green box indicates that the machine learning object detector has reached the desired level of confidence to identify a drone in that location, to be explained in more detail in the following paragraphs.

\begin{figure}[!htb]
    \centering
    \includegraphics[width=0.7\linewidth]{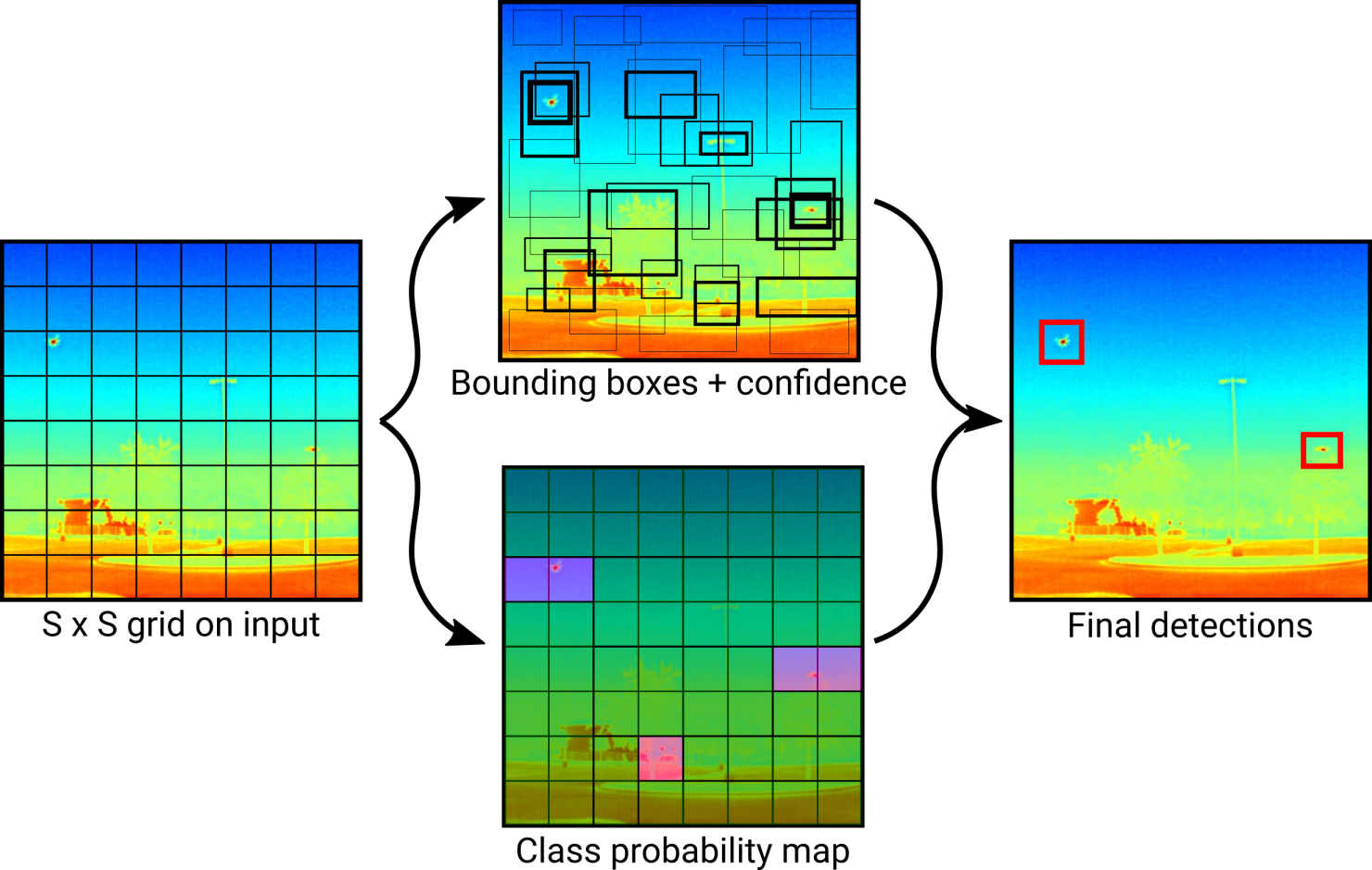}
    \caption{Illustration of the YOLO method for object detection. Adapted from original YOLO diagram \cite{redmon2016you}.}
    \label{fig:yolo_drone}
\end{figure}

In possession of a synchronized and annotated dataset of LWIR+RGB, LWIR, and RGB images, we trained a machine learning model for drone detection for each separate dataset using YOLOv3 \cite{redmon2013darknet,redmon2016you,redmon2018yolov3} official implementation\footnote{YOLOv3 on Github: \url{https://github.com/pjreddie/darknet}.}.
The LWIR and RGB models used their respective images and labels to train the YOLO object detector while the LWIR+RGB model used the combined frames, as seen in the leftmost image of Figure \ref{fig:sample_blended}, and all LWIR and RGB labels.
As illustrated in Figure \ref{fig:yolo_drone}, YOLO splits the input image in $S \times S$ grid cells, predicting one object and a fixed number of bounding boxes per cell using dimension clusters as anchor boxes \cite{redmon2017yolo9000}.
Each prediction of a bounding box is accompanied by a box confidence score and each grid cell by a conditional class probability, which, in our case, represents the probability of containing or not a drone.
These scores are condensed with a dot product in a final class confidence score that measures the confidence in both the classification (correct class of containing or not a drone) and the localization of the object (coordinates of the corners of the bounding box surrounding the drone).
YOLO was initially designed to be used with RGB images but no modification in the architecture was required since the LWIR+RGB frames still have 3 image channels, similar to an RGB image. The same applies to the LWIR frames, displayed in colors, which also have 3 image channels.
Due to the modularity of this approach, it can be integrated into existing surveillance and camera systems capable of recording LWIR and RGB data.

% ----------------------------------------------------
% RESULTS
% ----------------------------------------------------
\section{Results and Discussion}

The proposed approach is evaluated in four drone detection scenarios: drones crossing above and below the horizon and treeline, drones flying in front of cameras pointed directly the sun during sunrise, drones flying together with other heat sources such as public street lighting, and drone detection at approximately 300 meters away, here called long-range since the sUAS is barely visible on the RGB spectrum and also to differentiate of the previous cases where the vehicles were flying at less than 300 meters away.
We compared our approach of combining LWIR and RGB images and detecting drones using a custom trained YOLO model against the same YOLO architecture but trained only using LWIR and RGB images alone. All YOLO models for the baselines and the proposed method were trained on the same scenarios and using the same labels, only changing the input video appropriately.
Performance is computed in terms of \textit{Detection Rate} (DR) and \textit{False Alarm Rate} (FAR) \cite{Kim_2005}. DR is computed as the ratio between true positives and the sum of true positives (TP) and false negatives (FN), as seen in Equation \ref{eq:dr}, and FAR is computed as the ratio between false positives (FP) and the sum of true and false positives, as seen in Equation \ref{eq:far}.
Since YOLO is a probabilistic model, it outputs a confidence metric for each object detected in the image frame. In this work, we evaluate the performance metrics DR and FAR for all baselines and the proposed method for four confidence thresholds: $25$, $50$, $75$, and $90\%$.
Each specific condition is measured, explained, and illustrated bellow and qualitative results can be seen in the complete prediction videos for each condition shown in the project webpage\footnote{Project page: \url{https://sites.google.com/view/tamudrone-spie2020/}.}. 

% DR
\begin{equation} \label{eq:dr}
    DR = \frac{TP}{TP+FN}
\end{equation}

% FAR
\begin{equation} \label{eq:far}
    FAR = \frac{FP}{TP+FP}
\end{equation}

\begin{figure}[!htb]
    \centering
    \includegraphics[width=0.985\linewidth]{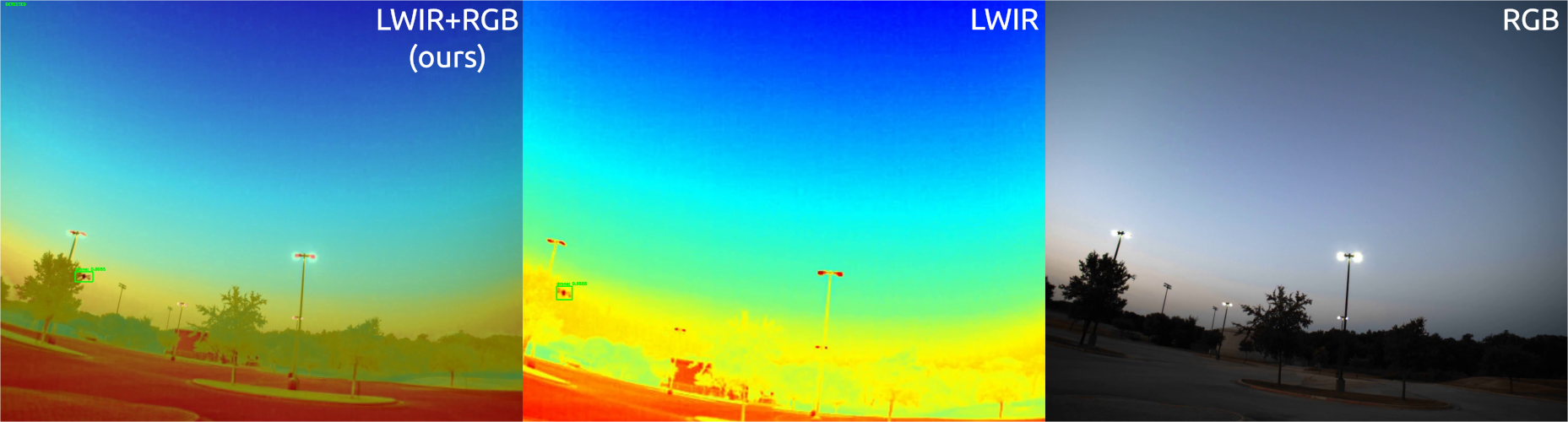}
    \caption{Frames illustrating our proposed approach (LWIR+RGB, leftmost image) compared to LWIR and RGB alone to detect sUAS when it is not visible on the RGB spectrum as, for example, when flying below the horizon in front of trees.}
    \label{fig:treeline}
\end{figure}

Our first experiment, shown in Figure \ref{fig:treeline}, has a drone crossing above and below the treeline and horizon.
This scenario is problematic for solutions that solely rely on the visible spectrum because of the drone’s low contrast relative to the background.
The infrared sensor, shown in the middle of Figure \ref{fig:treeline}, has a significantly higher contrast compared to the RGB sensor, due to salient heat emission from the drone batteries and motors.
This allows the sensor to keep track of the drone even when it crosses above and below the skyline.
In comparison, our proposed method, LWIR+RGB, presents a better horizon contrast when compared to LWIR and RGB alone.

\begin{figure}[!htb]
    \centering
    \includegraphics[width=0.985\linewidth]{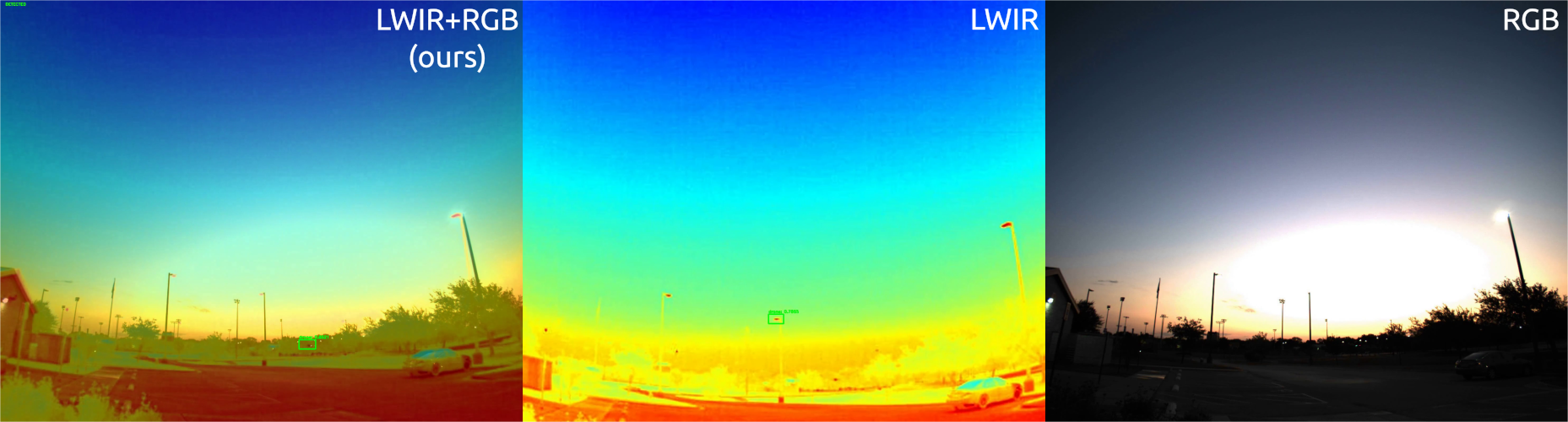}
    \caption{Frames illustrating our proposed approach (LWIR+RGB, leftmost image) compared to LWIR and RGB alone to detect sUAS when it is not visible on the RGB spectrum as, for example, during sunrise when cameras are pointed to the sun.}
    \label{fig:sunrise}
\end{figure}

Figure \ref{fig:sunrise} illustrates our second experiment with the drone flying in front of cameras pointed directly at the sun during sunrise.
This is another scenario where the RGB sensor does not have enough contrast for the proper detection of the sUAS.
Using both the visible and infrared spectrum, our solution can detect the drone with much less difficulty, while also rejecting the false positives that arise from purely relying on the infrared spectrum.

\begin{figure}[!htb]
    \centering
    \includegraphics[width=0.985\linewidth]{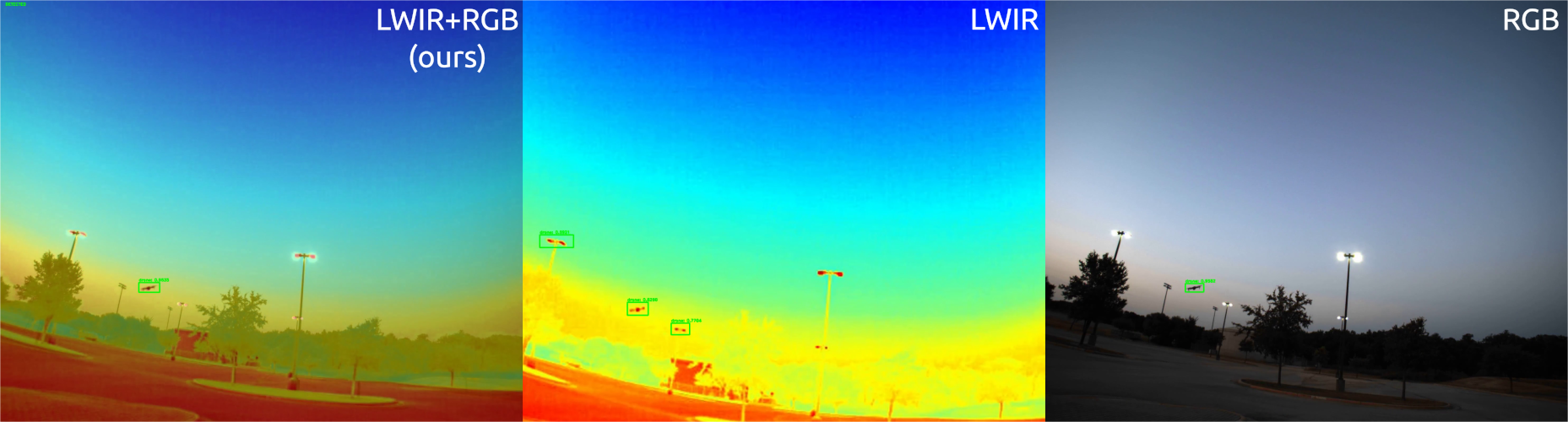}
    \caption{Frames illustrating our proposed approach (LWIR+RGB, leftmost image) compared to LWIR and RGB alone to detect sUAS when objects with similar heat signatures to drones, such as lamps, are present in the image.}
    \label{fig:birds}
\end{figure}

Our third experiment, shown in Figure \ref{fig:birds}, illustrates the solution’s robustness to other heat sources such as public street lighting.
This scenario is problematic for solutions that solely rely on the infrared spectrum because of the similarities of drone's and lamp's heat signatures at low resolution.
By combining both LWIR and RGB images, our method is still capable to leverage the heat signature of the drone to improve detection but also filters out light bulbs because they are clearly distinct from drones in the visible spectrum.
In this scenario, RGB alone still fails when the drones are below the horizon and in front of the trees.

\begin{figure}[!htb]
    \centering
    \includegraphics[width=0.985\linewidth]{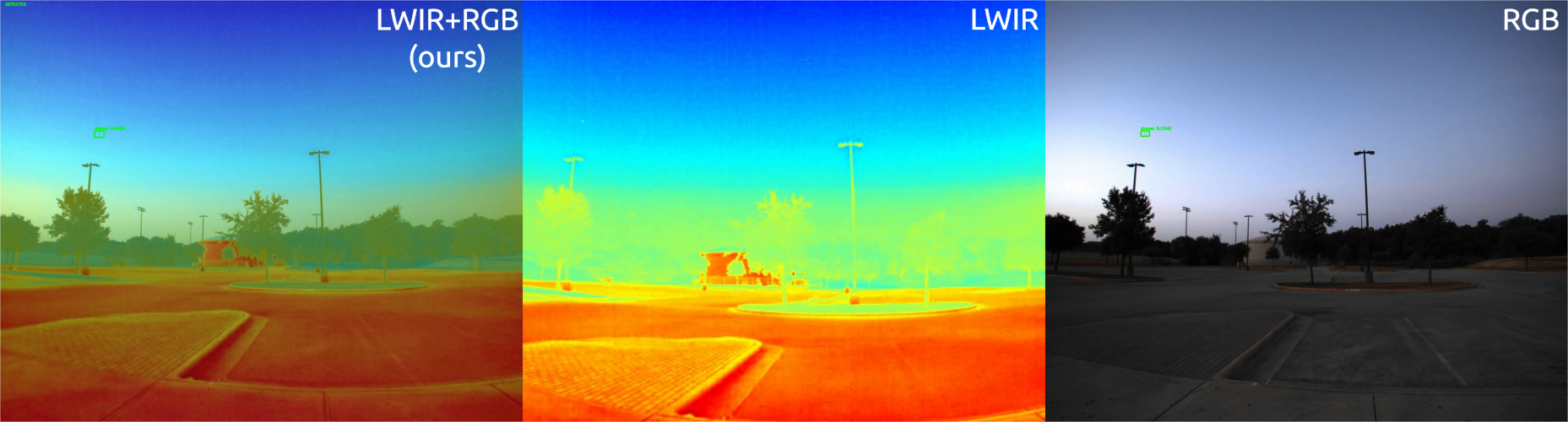}
    \caption{Frames illustrating our proposed approach (LWIR+RGB, leftmost image) compared to LWIR and RGB alone to detect sUAS flying beyond 300 meters from the cameras, here defined as long-range since the sUAS is barely visible on the RGB spectrum.}
    \label{fig:long}
\end{figure}

In our final experiment, shown in Figure \ref{fig:long}, we demonstrate our proposed method capabilities detecting drones when the sUAS is barely visible on the RGB spectrum, here called long-range to differentiate of the previous cases where the vehicles were flying at less than 300 meters away.
This scenario is challenging to LWIR alone due to the reduced resolution and still challenging for RGB alone depending on how the drone contrasts with the background.
As before, our proposed method still presents higher performance metrics for being able to combine both sensors, as explained in the following paragraph.

Table \ref{tab:dr_table} and \ref{tab:far_table} show the numerical results in terms of mean and standard deviation of Detection Rate (DR) and False Alarm Rate (FAR), respectively for each table, averaging the metrics for all scenarios and also showing them separately for cases where a single or multiple drones were flying at the same time, for all confidence threshold (CT) conditions of the machine learning model output.
In bold we highlighted the best performing method for each condition, showing that our method outperforms the baselines 87.5\% of the time.
Improvements in the DR ranges from 12.9 to 880.4\% while the reduction in FAR ranges from -15.4 to -100.0\%, depending on the scenario.
Overall, setting CT to $50\%$ presents a good balance of DR and FAR for all conditions investigated.
Qualitative results can be seen in the complete prediction videos for each condition shown in the project webpage\footnote{Project page: \url{https://sites.google.com/view/tamudrone-spie2020/}.}.

% DR TABLE WITH VARIATION
\begin{table}[!htb]
\centering
\caption{\textbf{Detection Rate (DR)} of our proposed method combining LWIR and visible spectrum imagery (LWIR+RGB) compared to using LWIR and the visible spectrum (RGB) alone. Results, in terms of the mean and standard deviation (SD), were computed for different scenarios containing single and multiple drones in the field of view of the cameras. Results are also controlled for the confidence threshold (CT) used to filter the outputs of the machine learning model. The ``LWIR+RGB Variation'' columns quantifies the difference between the baseline and our proposed method.}
\label{tab:dr_table}
\begin{tabular}{@{}lcccccc@{}}
\toprule
\multicolumn{1}{c}{\multirow{2}{*}{\textbf{Condition}}} & \multirow{2}{*}{\textbf{\begin{tabular}[c]{@{}c@{}}CT\\ (\%)\end{tabular}}} & \textbf{\begin{tabular}[c]{@{}c@{}}LWIR+RGB\\ (Ours)\end{tabular}} & \multicolumn{2}{c}{\textbf{LWIR}} & \multicolumn{2}{c}{\textbf{RGB}} \\ \cmidrule(l){3-7} 
\multicolumn{1}{c}{} &  & Mean $\pm$ SD & Mean $\pm$ SD & \begin{tabular}[c]{@{}c@{}}LWIR+RGB\\Variation\end{tabular} & Mean $\pm$ SD & \begin{tabular}[c]{@{}c@{}}LWIR+RGB\\Variation\end{tabular} \\ \midrule
\multirow{4}{*}{Combined} & 25 & \textbf{0.821$ \pm$ 0.067} & 0.617 $\pm$ 0.489 & \textbf{33.0\%} & 0.633 $\pm$ 0.144 & \textbf{29.6\%} \\
 & 50 & \textbf{0.712$ \pm$ 0.083} & 0.421 $\pm$ 0.374 & \textbf{69.0\%} & 0.546 $\pm$ 0.151 & \textbf{30.4\%} \\
 & 75 & \textbf{0.522$ \pm$ 0.1} & 0.279 $\pm$ 0.249 & \textbf{86.7\%} & 0.402 $\pm$ 0.136 & \textbf{29.7\%} \\
 & 90 & \textbf{0.279$ \pm$ 0.161} & 0.137 $\pm$ 0.137 & \textbf{103.0\%} & 0.2 $\pm$ 0.131 & \textbf{39.8\%} \\ \midrule
\multirow{4}{*}{\begin{tabular}[c]{@{}l@{}}Single\\ Drone\end{tabular}} & 25 & 0.849$ \pm$ 0.076 & \textbf{0.973 $\pm$ 0.024} & -12.8\% & 0.644 $\pm$ 0.2 & 31.8\% \\
 & 50 & \textbf{0.742$ \pm$ 0.1} & 0.657 $\pm$ 0.266 & \textbf{12.9\%} & 0.578 $\pm$ 0.2 & \textbf{28.3\%} \\
 & 75 & \textbf{0.57$ \pm$ 0.093} & 0.43 $\pm$ 0.196 & \textbf{32.4\%} & 0.436 $\pm$ 0.159 & \textbf{30.7\%} \\
 & 90 & \textbf{0.385$ \pm$ 0.066} & 0.203 $\pm$ 0.146 & \textbf{89.9\%} & 0.276 $\pm$ 0.109 & \textbf{39.5\%} \\ \midrule
\multirow{4}{*}{\begin{tabular}[c]{@{}l@{}}Multiple\\ Drones\end{tabular}} & 25 & \textbf{0.778$ \pm$ 0.025} & 0.082 $\pm$ 0.05 & \textbf{842.6\%} & 0.616 $\pm$ 0.042 & \textbf{26.2\%} \\
 & 50 & \textbf{0.667$ \pm$ 0.029} & 0.068 $\pm$ 0.044 & \textbf{880.4\%} & 0.498 $\pm$ 0.061 & \textbf{34.0\%} \\
 & 75 & \textbf{0.449$ \pm$ 0.069} & 0.052 $\pm$ 0.036 & \textbf{751.2\%} & 0.351 $\pm$ 0.124 & \textbf{27.7\%} \\
 & 90 & \textbf{0.121$ \pm$ 0.104} & 0.04 $\pm$ 0.026 & \textbf{202.9\%} & 0.085 $\pm$ 0.035 & \textbf{41.3\%} \\ \bottomrule
\end{tabular}
\end{table}

% FAR TABLE WITH VARIATION
\begin{table}[!htb]
\vspace*{1.5 cm} % space between tables to improve readability
\centering
\caption{\textbf{False Alarm Rate (FAR)} of our proposed method combining LWIR and visible spectrum imagery (LWIR+RGB) compared to using LWIR and the visible spectrum (RGB) alone. Results, in terms of mean and standard deviation (SD), were computed for different scenarios containing single and multiple drones in the field of view of the cameras. Results are also controlled for the confidence threshold (CT) used to filter the outputs of the machine learning model. The ``LWIR+RGB Variation'' columns quantifies the difference between the baseline and our proposed method.}
\label{tab:far_table}
\begin{tabular}{@{}lcccccc@{}}
\toprule
\multicolumn{1}{c}{\multirow{2}{*}{\textbf{Condition}}} & \multirow{2}{*}{\textbf{\begin{tabular}[c]{@{}c@{}}CT\\ (\%)\end{tabular}}} & \textbf{\begin{tabular}[c]{@{}c@{}}LWIR+RGB\\ (Ours)\end{tabular}} & \multicolumn{2}{c}{\textbf{LWIR}} & \multicolumn{2}{c}{\textbf{RGB}} \\ \cmidrule(l){3-7} 
\multicolumn{1}{c}{} &  & Mean $\pm$ SD & Mean $\pm$ SD & \begin{tabular}[c]{@{}c@{}}LWIR+RGB\\Variation\end{tabular} & Mean $\pm$ SD & \begin{tabular}[c]{@{}c@{}}LWIR+RGB\\Variation\end{tabular} \\ \midrule
\multirow{4}{*}{Combined} & 25 & \textbf{0.059 $\pm$ 0.043} & 0.18 $\pm$ 0.124 & \textbf{-66.7\%} & 0.08 $\pm$ 0.047 & \textbf{-25.4\%} \\
 & 50 & \textbf{0.027 $\pm$ 0.026} & 0.107 $\pm$ 0.075 & \textbf{-74.1\%} & 0.052 $\pm$ 0.041 & \textbf{-47.1\%} \\
 & 75 & \textbf{0.009 $\pm$ 0.009} & 0.062 $\pm$ 0.065 & \textbf{-85.1\%} & 0.026 $\pm$ 0.03 & \textbf{-64.4\%} \\
 & 90 & \textbf{0 $\pm$ 0.001} & 0.031 $\pm$ 0.031 & \textbf{-97.2\%} & 0.018 $\pm$ 0.025 & \textbf{-95.2\%} \\ \midrule
\multirow{4}{*}{\begin{tabular}[c]{@{}l@{}}Single\\ Drone\end{tabular}} & 25 & \textbf{0.041 $\pm$ 0.04} & 0.248 $\pm$ 0.106 & \textbf{-83.2\%} & 0.049 $\pm$ 0.028 & \textbf{-15.4\%} \\
 & 50 & \textbf{0.018 $\pm$ 0.025} & 0.135 $\pm$ 0.077 & \textbf{-86.2\%} & 0.023 $\pm$ 0.014 & \textbf{-19.4\%} \\
 & 75 & 0.006 $\pm$ 0.009 & 0.073 $\pm$ 0.083 & -90.9\% & \textbf{0.004 $\pm$ 0.002} & 46.3\% \\
 & 90 & \textbf{0.001 $\pm$ 0.002} & 0.032 $\pm$ 0.04 & \textbf{-95.6\%} & 0.001 $\pm$ 0.002 & \textbf{-25.2\%} \\ \midrule
\multirow{4}{*}{\begin{tabular}[c]{@{}l@{}}Multiple\\ Drones\end{tabular}} & 25 & 0.087 $\pm$ 0.043 & \textbf{0.077 $\pm$ 0.063} & 12.3\% & 0.126 $\pm$ 0.019 & -31.2\% \\
 & 50 & \textbf{0.041 $\pm$ 0.029} & 0.065 $\pm$ 0.068 & \textbf{-36.5\%} & 0.096 $\pm$ 0.013 & \textbf{-57.1\%} \\
 & 75 & \textbf{0.013 $\pm$ 0.01} & 0.047 $\pm$ 0.05 & \textbf{-71.6\%} & 0.058 $\pm$ 0.007 & \textbf{-77.3\%} \\
 & 90 & \textbf{0 $\pm$ 0} & 0.029 $\pm$ 0.027 & \textbf{-100.0\%} & 0.042 $\pm$ 0.026 & \textbf{-100.0\%} \\ \bottomrule
\end{tabular}
\end{table}

% ----------------------------------------------------
% CONCLUSION
% ----------------------------------------------------
% \newpage
\section{Conclusion}

In this paper, we have demonstrated that using state-of-the-art deep learning algorithms for object detection with the insight of combining long-wave infrared (LWIR) and visible spectrum (RGB) sensors can significantly improve the detection of sUAS in previously difficult situations as, for example, when sUAS flies crossing above and below the treeline and horizon line, in the presence of other heat sources, with cameras pointed directly to the sun, and in long-range scenarios.
We compared our approach of combining LWIR and RGB images and detecting drones using a custom trained YOLO model against the same YOLO architecture but trained only using LWIR and RGB images alone.
Our approach achieved a detection rate of 71.2 $\pm$ 8.3\%, improving by 69\% when compared to LWIR and by 30.4\% when visible spectrum alone, and achieved false alarm rate of 2.7 $\pm$ 2.6\%, decreasing by 74.1\% and by 47.1\% when compared to LWIR and visible spectrum alone, respectively, on average, for single and multiple drone scenarios, controlled for the same confidence metric of the machine learning object detector of at least 50\%.
We believe that the further development of this technology has serious benefits to the area of counter-sUAS and can supplement existing technology in practical applications and enable counter-sUAS systems to operate in a 24/7 continuous regime and prevent malicious use of the sUAS.

% ----------------------------------------------------
% LIMITATIONS AND FUTURE WORK
% ----------------------------------------------------
\section{Limitations and Future Work}

There are still some inherent limitations to the vision-based approaches for object detection.
One of those limitations stems from scenarios where the drone is indistinguishable from objects with similar visual characteristics and heat signatures.
Although the multi-modal approach mitigated this issue, the system had difficulty distinguishing the drone from the birds during long-range detection since both objects have a similar heat signatures at low resolution.
Besides using higher-resolution sensors and collecting more training data, one avenue we are exploring to further mitigate this limitation uses a filtering algorithm, like Deep SORT \cite{Bewley2016_sort}, to track drones and other objects over the individual frames.
From there, each object's flight characteristics are used as another input for detection and provide a more educated estimate for indiscernible heat signatures. 
In combination with other sensor modalities like a radar and spectrum analyzer, the solution presented could offer an accurate and robust solution to sUAS detection.

% \appendix    %>>>> this command starts appendixes

\acknowledgments % equivalent to \section*{ACKNOWLEDGMENTS}       
 
The authors would like to thank every other participant of the ``A-Team", winning team of the A-Hack-of-the-Drones 2018, where this project started: Edan Coben, Emily Fojtik, Garrett Jares, Humberto Ramos, Nicholas Waytowich, Niladri Das, Sunsoo Kim, Vedang Deshpande, Venkata Tadiparthi, and Vernon Lawhern; Texas A\&M University Professors: Dr. Daniel Ragsdale, Dr. John Hurtado, and Dr. Raktim Bhattacharya, for supporting the students in the event; Rodney Boehm from Texas A\&M University Engineering Entrepreneurship program for supporting the work after the event; and MD5 and Army Future Commands for sponsoring the event and following up with the team to continue the work after the hackathon.

% References
\bibliography{refs} % bibliography data in report.bib
\bibliographystyle{spiebib} % makes bibtex use spiebib.bst

\end{document}